\documentclass[graybox]{svmult}

\usepackage{mathptmx}       % selects Times Roman as basic font
\usepackage{helvet}         % selects Helvetica as sans-serif font
\usepackage{courier}        % selects Courier as typewriter font
\usepackage{type1cm}        % activate if the above 3 fonts are
\usepackage{makeidx}         % allows index generation
\usepackage{graphicx}        % standard LaTeX graphics tool
\usepackage{multicol}        % used for the two-column index
\usepackage[bottom]{footmisc}% places footnotes at page bottom

\usepackage{graphics} % for pdf, bitmapped graphics files
\usepackage{epsfig} % for postscript graphics files
\usepackage{mathptmx} % assumes new font selection scheme installed
\usepackage{times} % assumes new font selection scheme installed
\usepackage{amsmath} % assumes amsmath package installed
\usepackage{amssymb}  % assumes amsmath package installed
\usepackage{bm}  % add by xingchao

\usepackage{subfigure}

\usepackage{wrapfig}

\makeindex             % used for the subject index
\begin{document}

\title*{Adapting control policies from simulation to reality using a pairwise loss}

\titlerunning{Adapting control policies from simulation to reality}
\author{Ulrich Viereck, Xingchao Peng, Kate Saenko, and Robert Platt}

\institute{Ulrich Viereck, Robert Platt \at College of Computer and Information Science, Northeastern University, \email{\{uliv,rplatt\}@ccs.neu.edu}
\and Xingchao Peng, Kate Saenko \at Department of Computer Science, Boston University, \email{\{xpeng,saenko\}@bu.edu}}
\maketitle

\vspace{-1.25in}

\abstract{This paper proposes an approach to domain transfer based on a pairwise loss function that helps transfer control policies learned in simulation onto a real robot. We explore the idea in the context of a ``category level'' manipulation task where a control policy is learned that enables a robot to perform a mating task involving novel objects. We explore the case where depth images are used as the main form of sensor input. Our experimental results demonstrate that proposed method consistently outperforms baseline methods that train only in simulation or that combine real and simulated data in a naive way.}

%%%%%%%%%%%%%%%%%%%%%%%%%%%%%%%%%%%%%%%%%%%%%%%%%%%%%%%%%%%%%%%%%%%%%%%%%%%%%%

\section{Motivation and Problem Statement}
\label{sec:1}
% (1 page)

\begin{wrapfigure}{r}{0.5\textwidth}
    \centering
    \vspace{-0.3in}
    \subfigure{\includegraphics[width=0.5\textwidth]{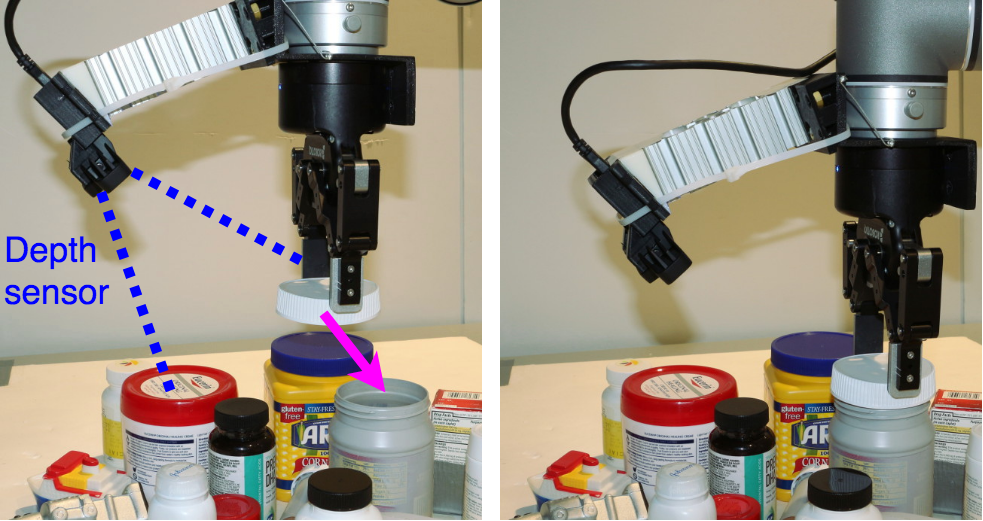}}
    \vspace{-0.1in}    
    \caption{Our goal is to learn a controller that uses depth image feedback to mate the cap to the bottle in the presence of clutter. Experimental setup on UR5 robot with a Intel RealSense depth sensor mounted as shown.}
    \label{UR5_setup}
    \vspace{-0.1in}
\end{wrapfigure}

Recently, there has been a lot of interest in using deep neural networks to learn ``pixels-to-torques'' visuomotor controllers: robotic controllers that take sequential image data as input and produce low level motor commands as output. Ideally, we would learn these controllers with training data collected using real robotic hardware~\cite{levine2016learning}. However, this approach is rarely feasible because of the large amounts of training experience typically required to train deep neural networks. Instead, it is convenient to learn pixels-to-torques control policies using simulated data. 

Unfortunately, this exposes us to the \textit{dataset shift problem}~\cite{dabook09}. When the simulation is not sufficiently accurate, then the control policy learned in simulation may not work well in reality. There are two types of simulation errors that typically can cause domain shift problems: 1) errors simulating images that the robot would observe, and 2) errors simulating real world contact and frictional dynamics. This paper limits consideration to non-contact tasks and therefore the focus is on domain shift errors caused by image simulation and depth image simulation in particular.

In this paper, we propose an approach to transferring visuomotor control policies learned on simulated depth data to real world observations. The key idea is to reduce the gap between simulation and reality by augmenting the simulated data used to train the system with a small amount of real robot data. Each piece of real robot data is paired with a piece of simulated data that corresponds to the same robot state. We train the neural network using a loss function that has two terms: a \textit{task loss} that encodes the desired robotic behavior and a \textit{pairwise loss} that penalizes networks that do not represent real and simulated data the same way. 

We make two contributions relative to prior work: 1) we propose a neural network architecture that combines the pairwise loss approach to domain transfer with a pixels-to-torques controller; 2) we characterize the method for depth image data rather than for RGB data. Unlike~\cite{tzeng2015adapting} which uses the pairwise loss function as part of the state estimator and only explores single task instances, our approach learns controllers that can solve ``category level'' manipulation tasks. We find that the approach can work well even when the real data is produced in a simplified version of the actual robotic scenario that is experienced at test time.

\section{Related Work}
\label{sec:related}
This paper complements a variety of recent literature on simulation-to-reality learning for robotics tasks. One example is recent work that uses ``domain randomization'' of simulated images to affect better transfer to reality~\cite{tobin2017domain}. Another example is work using GANs to affect the transfer~\cite{bousmalis2017using}. In contrast, the approach followed here is simpler than GAN-based approaches and more relevant to depth data than domain randomization methods.

The most related approach to ours is that of~\cite{tzeng2015adapting}. It also uses the pairwise loss function to minimize domain shift between simulated and real data. However, it only explores single task instances (e.g., placing a specific rope object on a specific scale object), whereas we solve ``category level'' manipulation tasks where object shape and size varies from one instance of the task to the next. Furthermore,~\cite{tzeng2015adapting} uses keypoint prediction on RGB image as an intermediate state representation, whereas we estimate a distance-to-goal function directly. We also use depth data which tends to be less affected by domain shift due to lighting and background than RGB data. 
%%%%%%%%%%%%%%%%%%%%%%%%%%%%%%%%%%%%%%%%%%%%%%%%%%%%%%%%%%%%%%%%%%%%%%%%%%%%%%

\section{Technical Approach}
\label{sec:2}

\begin{figure}[tb]
    \centering
    \includegraphics[width=0.7\textwidth]{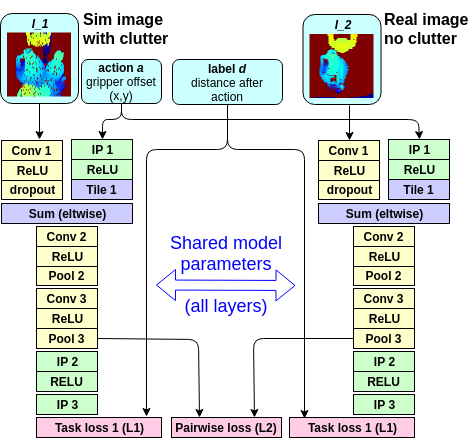}
    \caption{Architecture of the controller neural network. The pairwise loss at pool3 favors networks that give real and simulated images similar representations.}
    \label{fig:CNN}
\end{figure}

\subsection{Controller Network Architecture}

We learn a pixels-to-torques controller that takes depth images as input and outputs manipulator displacements. The controller is based on a method proposed in our prior work~\cite{pmlr-v78-viereck17a} where we estimate a distance function with respect to a goal  state using a neural network. Given an image and a candidate manipulator displacement, the distance function predicts expected distance-to-goal on the following time step. We select a manipulator displacement by sampling a set of candidate displacements and selecting the one that is predicted to move closest to a goal. Essentially, this method learns a value function over the cross product of observation and action (depth image and manipulator displacement). However, instead of using reinforcement learning, we train the neural network directly by using supervised learning with distance targets produced by our simulator. Specifically, we create a dataset by sampling from a space of possible task scenarios and initial conditions. For each sample, we simulate the depth image that would be observed and calculate distance-to-goal after performing the associated displacement. 

\subsection{Supervised Domain Adaptation}
This paper characterizes a domain transfer technique based on the following loss function (similar to what was originally proposed in~\cite{tzeng2015adapting}):
\begin{eqnarray}
\nonumber
\mathcal{L} & = & \alpha \sum_{(I,a) \in \mathbb{X}_S} \| g \left( f(I,a; \theta_f), \theta_g \right) - y(I,a) \|_1 \\
\nonumber
& + & \beta \sum_{(I,a) \in \mathbb{X}_T} \| g \left( f(I,a; \theta_f), \theta_g \right) - y(I,a) \|_1 \\
& + & \gamma \sum_{(I_S,I_T,a) \in \mathbb{X}_{ST}} \Arrowvert f(I_S,a;\theta_f) - f(I_T,a;\theta_f) \Arrowvert_2.
\label{eqn:loss}
\end{eqnarray}

\noindent
Here, we write the neural network as a composition of two functions, $f$ and $g$. $f$ denotes the ``early'' part of the network comprised of convolutional layers that learn feature representations for the images and actions. $g$ denotes the ``late'' fully connected layers that encode the regression task. The output of the end-to-end network for an image $I$ and action (i.e. displacement) $a$ is written $g(f(I,a;\theta_f);\theta_g)$, where $\theta_f$ denotes the parameters of $f$ and $\theta_g$ denotes parameters of $g$. The loss is evaluated over a training dataset comprised of two parts: the data from source domain and the data from the target domain. The source domain consists of simulated images and actions $(I,a) \in \mathbb{X}_S$ and the associated labels $y(I,a)$ (these are the distance-to-goal that would result from taking action $a$ from the state that produced image $I$). The target domain $\mathbb{X}_T$ consists of real images and actions paired with the associated labels. The set $\mathbb{X}_{ST}$ consists of triples $(I_S,I_T,a)$ where $I_S$ and $I_T$ are the simulated-real image pair and $a$ is the action that was taken. We assume that the cardinality of $\mathbb{X}_S$ is much larger than either $\mathbb{X}_{T}$ or $\mathbb{X}_{ST}$, i.e. that we have many more simulated images than real images. The first two terms of Equation~\ref{eqn:loss} are called \textit{task losses}. The third is the \textit{pairwise loss}. The first and second task losses are minimized when the neural network fits the simulated and real data well, respectively. The third is minimized when the network assigns both simulated and real depth images the same high-level encoding. The pairwise loss is critical: by ``preferring'' networks that encode real and simulated data similarly, this term facilitates good transfer from simulation to reality. The approach is implemented by the neural network architecture illustrated in Figure~\ref{fig:CNN}. It takes as input a pair of depth images, $I_1$ and $I_2$ (matching paired images from simulation and reality), and an action $a = (x,y) \in \mathbb{R}^2$. It learns a function, $d(I,a) \in \mathbb{R}_{>0}$, that describes the distance between the object in the hand and the target after displacing the manipulator by $a$.

\subsection{Data Simulation}
We train using a combination of real and simulated data. The simulated portion of the dataset is generated using OpenRAVE~\cite{openrave_website} to generate 100k 64x64 pixel depth images. Each depth image is taken for a random robot configuration with clutter objects selected randomly from a set of more than 250 objects and placed randomly in the vicinity of the target bottle. For each of the 100k scenes, we sample 100 actions (i.e. hand displacements) and estimate the distance-to-goal that would result if that action were executed. Finally, we simulate missing pixel noise by setting the value of each pixel to 0 with a 10\% probability. We also collect 7260 labeled real training images on the robot (UR5, see Figure~\ref{UR5_setup}) and measure the corresponding ground truth distance-to-goal. Importantly, the real images used for training do \textit{not} have clutter. As a result, it is easier to obtain training data semi-automatically because it is unnecessary to reproduce simulated clutter on the real system. These real images were paired with 7260 simulated images that portray the same robotic state.

\subsection{Unsupervised Domain Adaptation} 
One alternative way to tackle the simulated-to-real knowledge transfer is to leverage unsupervised domain adaptation algorithms~\cite{long2015learning}. Assume two distributions $p$ and $q$ are sampled from simulated and real domain, respectively, we apply Maximum Mean Discrepancy (MMD)~\cite{gretton2012kernel} to align $p$ and $q$. Denoting the reproducing kernel Hilbert space (RKHS) with a characteristic kernel $k$ by $\mathbb{H}_k$. The MMD $d_k(p,q)$ is defined as

\begin{equation}
     d^2_k(p,q) \overset{\Delta}{=} \| \bm{E}_p[f(I_S)] - \bm{E}_q[f(I_T)] \|_{\mathbb{H}_k}^2
\label{eq_mmd} 
\end{equation}

\noindent where $I_S$ and $I_T$ are unpaired images from source (simulated) and target (real) domains. The main limitation of the algorithm is the computation complexity is quadratic. In this paper, following~\cite{gretton2012optimal}\cite{long2015learning}, we adopt the unbiased estimation of MMD which can be computed with cost $O(n)$. Specifically, 

\begin{equation}
\begin{aligned}
     d^2_k(p,q) \overset{\Delta}{=} & \frac{2}{n_s}\sum_{i=1}^{n_s/2} [k(f(I_S^{2i-1}), f(I_S^{2i})) + k(f(I_T^{2i-1}), f(I_T^{2i}))\\ 
     & -k(f(I^{2i-1}_S), f(I^{2i}_T)) - k(f(I^{2i}_S), f(I^{2i-1}_T))]
\end{aligned}
\label{eq_mmd_est} 
\end{equation}
\noindent where $n_s$ is the number of instances in source domain. The training loss with unsupervised domain adaptation is:

\begin{eqnarray}
\nonumber
\mathcal{L} & = & \alpha \sum_{(I,a) \in \mathbb{X}_S} \| g \left( f(I,a; \theta_f), \theta_g \right) - y(I,a) \|_1 \\
& + & \gamma  d^2_k(p,q).
\label{eqn_loss_with_da}
\end{eqnarray}
   
\section{Experiments}
\label{sec:3}

\begin{figure}
    \centering
    \subfigure{\includegraphics[width=0.8\textwidth]{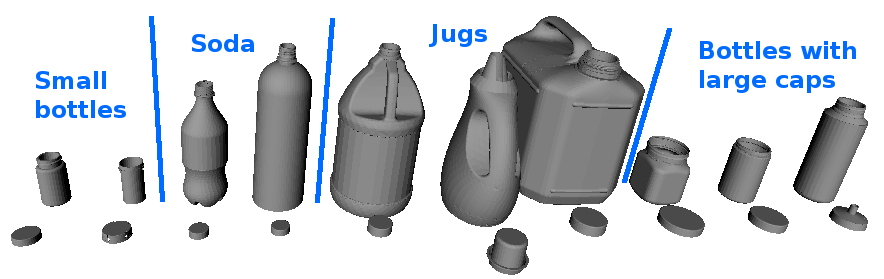}}\\
    \subfigure{\includegraphics[width=0.8\textwidth]{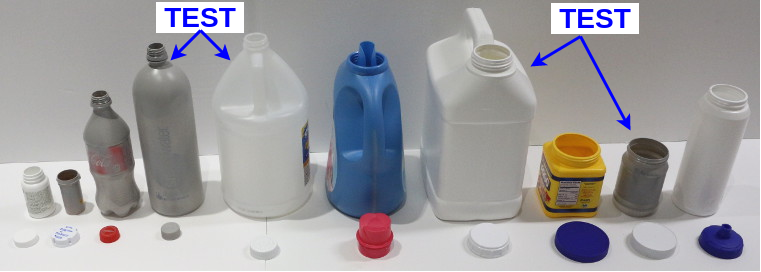}}%\hspace{0.05in}    
    \caption{Simulated and real bottles used in our experiments. From left to right: \textit{vitamin, pills, soda, water, 1-gal, laundry, 2-gal, corn starch, peanut butter, sport bottle}. Test bottles are shown with arrows.}
    \label{fig:sim_and_real_bottles}
\end{figure}

\subsection{Task Description}
We evaluate this approach using the \emph{cap-on-bottle} task, where the robot must align a cap with a bottle opening. This is a challenging task because we are learning a ``category level'' manipulation skill where the system must learn to perform the task for novel object instances in the presence of randomly placed clutter.  We test our methods in simulation, as well as on a real robot using a Robotiq 85 two-fingered hand mounted on a UR5 arm. The Intel RealSense SR300 is mounted to the UR5 wrist as shown in Figure~\ref{UR5_setup}. The RealSense creates depth images that are input to our controller and used to estimate desired hand displacements. 

 Testing on the robot proceeds as follows. A test bottle is placed on a table surrounded by clutter intended to make the task more challenging. Most of the clutter objects are bottles \emph{with} caps, which makes the task even harder. The bottle cap is manually placed into the robotic hand. The position of the bottle is unknown to the algorithm but measured for the purposes of experimental evaluation. The gripper is initialized to a random offset within a 10cm box centered on the bottle at a height of 5cm above the table. At each iteration, the controller acquires a depth image, samples 1k manipulator displacements, moves the gripper in the direction predicted to reduce distance-to-goal by the most, and moves toward the table by 1cm. After executing 5 iterations, we measure how close the cap is to the bottle opening.

\begin{figure}[tb]
    \centering
    \subfigure{\includegraphics[width=\textwidth]{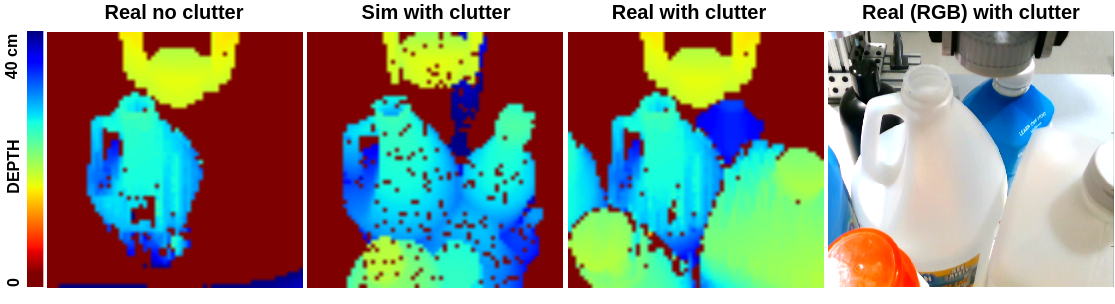}}
    \caption{Depth images generated for training and testing. For training \textit{Real no clutter} are paired and augmented by \textit{Sim with clutter} images. Testing images are \textit{Real with clutter} images. All these images show the same 1-gal bottle for comparison of different image domains, but we do not test on bottles that are in the training set (see Section~\ref{sec:3} for details). The \textit{Real (RGB) with clutter} images are not used (reference only).}
    \label{sim_real_rgb}
\end{figure}

\subsection{Experimental Settings}
Our proposed approach is to train the neural network using both the task and pairwise loss (supervised adaptation) over both simulated and real images. We compare this approach against five alternative setups: 1) training the network using only real images \textit{without} clutter; 2) training using only real images \textit{with} clutter; 3) training using simulated images with clutter; 4) training using simulated images with clutter with \emph{unsupervised adaptation} on real images \emph{with} clutter (MMD method) 5) training using simulated images with clutter combined with \emph{labeled} real images \emph{without} clutter. Among the five setups, pairwise losses are directly applied to layer \textit{pool3}. For the MMD method, we add a fully connected layer with 512 channels after layer \textit{conv1} and apply MMD loss on the top of the fully connected layer. We choose layer \textit{conv1} as it only contains image information, which makes more sense to align, as opposed to later layers which also contain pose information. We will refer to the combination of the five methods and the proposed approach as the six ``domain transfer methods''.

\begin{figure}[tb]
    \centering
    \includegraphics[width=\textwidth]{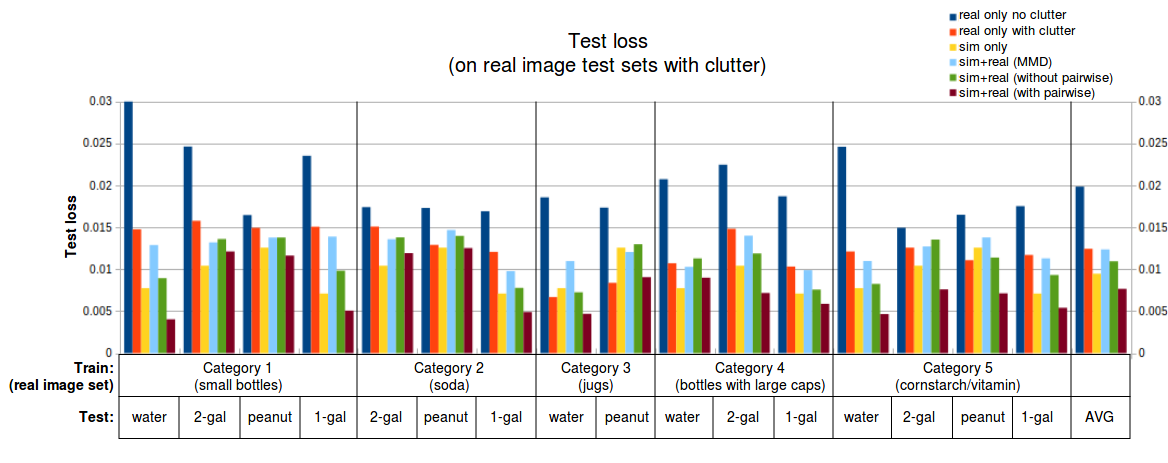}\vspace{-0.1in}
    \caption{Test losses on real image test sets with clutter. See text for details. }
    \label{fig:loss}
\end{figure}

\begin{figure}[tb]
    \centering
    \includegraphics[width=\textwidth]{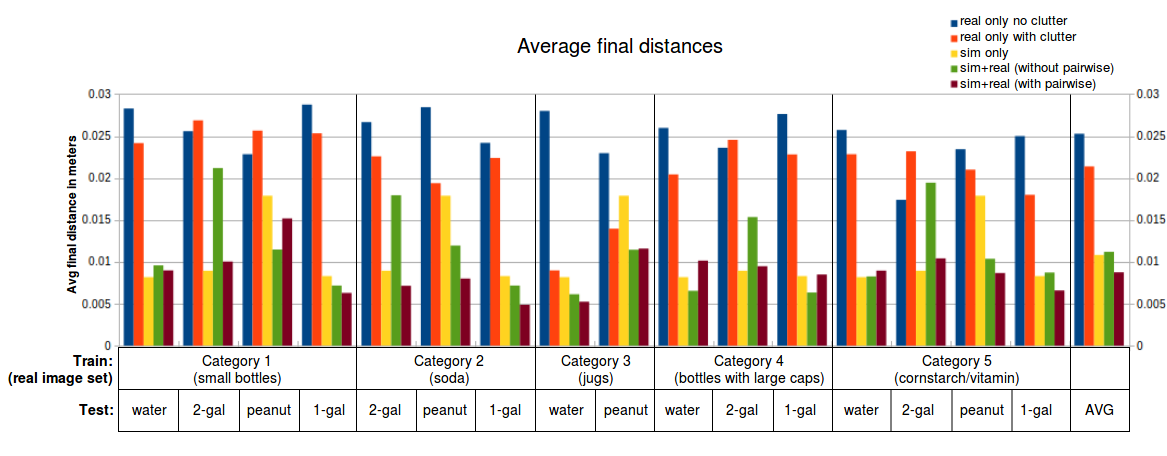}\vspace{-0.1in}
    \caption{Average final distances to cap on real image test sets with clutter. See text for details. }
    \label{fig:dist}
\end{figure}

\begin{figure}[tb]
    \centering
    \includegraphics[width=\textwidth]{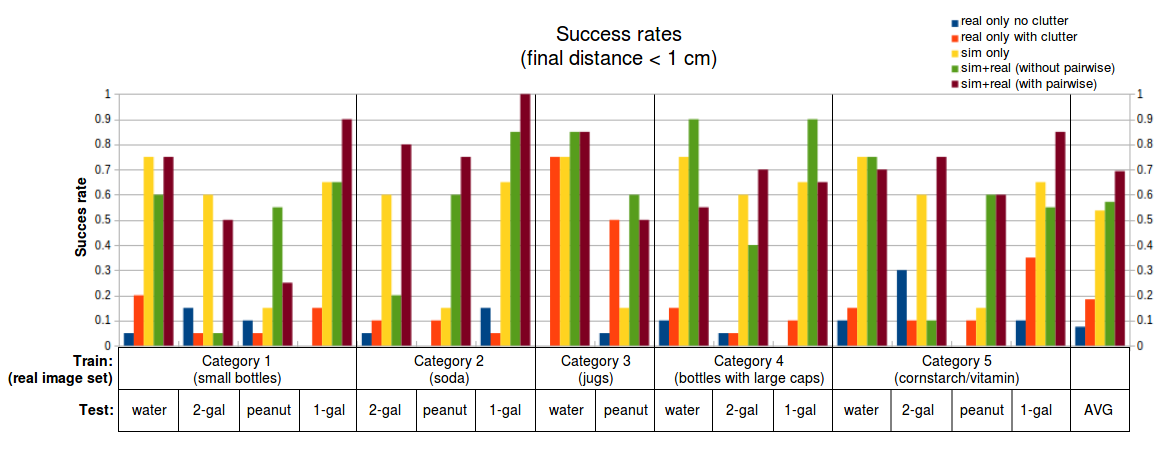}\vspace{-0.1in}
    \caption{Average success rates on real image test sets with clutter. See text for details.}
    \label{fig:success}
\end{figure}

We evaluate the six domain transfer methods over a set of five different problem scenarios. In each problem scenario, we train using data derived from a different object category. The five object categories are: (i) small pill bottles; (ii) soda bottles; (iii) jugs; (iv) bottles with large caps; (v) cornstarch and vitamin bottles. These categories are illustrated in Figure~\ref{fig:sim_and_real_bottles}.  A separate network is trained for each object category and is evaluated on as many as four different test objects (indicated in the figure with arrows). We do not test on objects that happen also to be in the object category used for training. 
\vspace{-0.2in}
\subsection{Evaluation}
Figures~\ref{fig:loss},~\ref{fig:dist},~\ref{fig:success} show the results grouped by object category. For example, the ``Category 1'' results show performance for each of the six domain transfer methods evaluated on four different test objects (water, 2-gal, peanut, and 1-gal). 
Figure~\ref{fig:loss} show L1 test loss of the learned network (lower is better).  Figure~\ref{fig:dist} shows the average final distance between cap and bottle opening after the last control iteration. Sometimes the controller diverged or the cap was moved to a distractor object. Therefore the final distance is set to 3 cm for final distances $>$ 3 cm to reduce the effect of diverged runs on the average. Each bar is an average of 20 trials (lower is better). Figure~\ref{fig:success} shows the success rate of the end-to-end controller (we define ``success'' to occur when the cap final position is within 1cm of the position of the bottle top). Each bar is an average of 20 trials (higher is better). Note that for the unsupervised domain adaptation baseline ``sim+real (MMD)" we have only results for the ``test loss", not for ``final distance" and ``success rates". However, comparing the three bar plots for the other domain transfer methods shows that the test loss roughly reflects the performance on the robot.\\
The baseline ``sim only" consists of 100k simulated images as described in Section~\ref{sec:2}. The dataset used for domain transfer method ``sim+real (MMD)" contains 100k labeled simulated images with 726 unlabeled real images with clutter per bottle in the category (e.g. 2x726 for category 2 ``soda" or 3x726 for category 3 ``jugs"). We use real images \emph{with} clutter, since this matches closely the distribution of the target domain (we test with clutter). However, our approach is able to use real images \emph{without} clutter with better performance. The datasets for ``sim+real (without pairwise)" and ``sim+real (with pairwise)" contain \emph{labeled} real images \emph{without} clutter paired with the corresponding simulated images in the same robot state. The real images in the dataset are repeated to obtain 100k real images to balance the 100k simulated images. To prevent overfitting to the real images, the weight for the task loss of the real images is set to 0.1 (the weight for for the task loss of the simulated images remains 1.0). The weight for the pairwise loss for ``sim+real (with pairwise)" is set to 0.1 and the weight for MMD loss for MMD baseline is set to 0.05 (weight loss parameters were determined by hyperparameter tuning).

In Figure~\ref{fig:loss}, we show the experimental results of the unsupervised domain adaptation method. The MMD-based method works better than ``real image with clutter,'' despite the fact that it does not use any supervision for the real images. However, it performs worse than ``sim+real without pairwise loss''. After some empirical analysis, we think it is mainly caused by three reasons. Firstly, the real data is unlabeled in the unsupervised domain adaptation setting. Secondly, the real images used for training and testing are also biased by themselves. Thirdly, the domain adaptation loss is not directly applied to align the action information while the test loss is highly relevant to that.

\begin{figure}[t]
    \subfigure{\includegraphics[width=\textwidth]{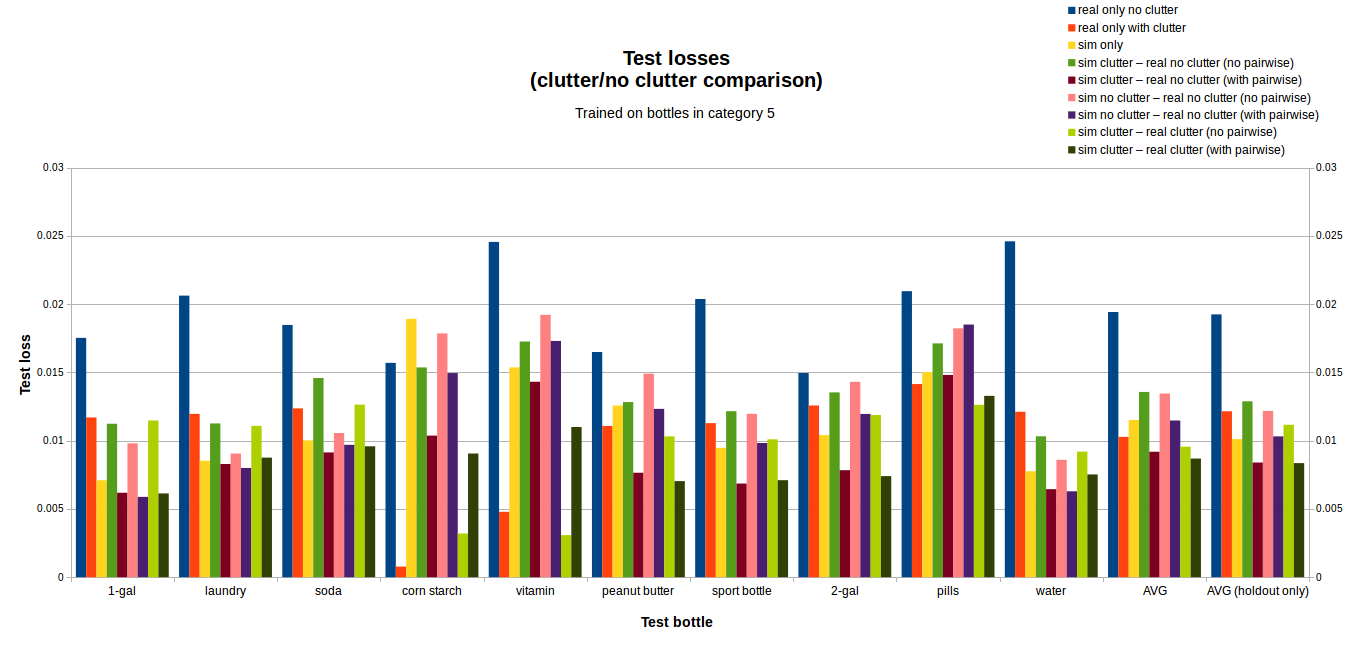}}\vspace{-0.2in}
    \caption{Effect of different clutter/no clutter combinations in the pairwise sim-real image dataset. This plot also shows the effect adding pairwise loss. Without pairwise loss the network tends to overfit to the real images. In particular the network trained on ``real only with clutter" and ``sim clutter - real clutter (no pairwise)" overfit to the two training bottles ``corn starch" and "vitamin bottle" in the category 5 bottles.}
    \label{fig:losses_for_clutter_no_clutter_comparison}
\end{figure}

We experimented with clutter and no clutter combinations in the sim-real image pairs. Figure~\ref{fig:losses_for_clutter_no_clutter_comparison} shows the test loss when trained on different combinations of having clutter in sim or real part of the image pairs. For this experiment we trained on bottles from category 5 (this category contains cornstarch and vitamin bottle) and tested on all available bottles. There are two averages across test bottles: ``AVG" is the average over all bottles (including bottles trained on) and ``AVG (holdout only)" excluding the training bottles. Note that the ``sim-real" datasets also contains 100k sim images \emph{with} clutter. The main insight of this experiment is that using sim image \emph{with clutter} and real image \emph{without} clutter gives almost as good of a performance as real image \emph{with} clutter in the case of applying pairwise loss. Also, the gap between ``with" and ``without" pairwise is the largest. Note how the nets containing \emph{real images with clutter} overfits to ``corn starch" and ``vitamin" if no pairwise loss is applied. This shows that the pairwise loss has a regularization effect and prevents overfitting to the training data.

%\vspace{-0.2in}

\section{Discussion and Conclusion}
The group of bars at the far right of Figures~\ref{fig:loss}-\ref{fig:success} labeled ``AVG'' summarize the comparison. Each bar (i.e. each domain transfer method) is an average of all experiments for that domain transfer type. This illustrates a few key results. First, the domain transfer method using our proposed task-pairwise loss function does best overall (lowest loss, lowest average final distance, highest task success rate). Second, the method using only real images does worst, probably because we do not train on enough different objects to facilitate generalization to novel objects. Third, training using only simulated images does quite well (~54\% success rate): not quite as well as we can do using our proposed method (~70\% success rate), but not nearly as badly as training on only real data.

The unsupervised domain adaptation approach performs worse than trained on ``sim only" (see Figure~\ref{fig:loss}). The reason might be that simulated and real depth images look very similar (they are both depth images with similar distributions of gray scale values) and the domain adaptation does not have much of an effect.

Overall, we conclude that training visuomotor policies for category-level tasks in simulation is a promising approach, and that by collecting a small amount of labeled real data in simplified scenarios and using the pairwise loss, we can improve performance on real systems.

\vspace{-0.25in}

\bibliographystyle{plain}

\bibliography{main}

\begin{thebibliography}{10}

\bibitem{bousmalis2017using}
Konstantinos Bousmalis, Alex Irpan, Paul Wohlhart, Yunfei Bai, Matthew Kelcey,
  Mrinal Kalakrishnan, Laura Downs, Julian Ibarz, Peter Pastor, Kurt Konolige,
  et~al.
\newblock Using simulation and domain adaptation to improve efficiency of deep
  robotic grasping.
\newblock {\em arXiv preprint arXiv:1709.07857}, 2017.

\bibitem{openrave_website}
R.~Diankov.
\newblock Openrave.
\newblock {\em http://openrave.org}.

\bibitem{gretton2012kernel}
Arthur Gretton, Karsten~M Borgwardt, Malte~J Rasch, Bernhard Sch{\"o}lkopf, and
  Alexander Smola.
\newblock A kernel two-sample test.
\newblock {\em Journal of Machine Learning Research}, 13(Mar):723--773, 2012.

\bibitem{gretton2012optimal}
Arthur Gretton, Dino Sejdinovic, Heiko Strathmann, Sivaraman Balakrishnan,
  Massimiliano Pontil, Kenji Fukumizu, and Bharath~K Sriperumbudur.
\newblock Optimal kernel choice for large-scale two-sample tests.
\newblock In {\em Advances in neural information processing systems}, pages
  1205--1213, 2012.

\bibitem{levine2016learning}
Sergey Levine, Peter Pastor, Alex Krizhevsky, and Deirdre Quillen.
\newblock Learning hand-eye coordination for robotic grasping with deep
  learning and large-scale data collection.
\newblock {\em arXiv preprint arXiv:1603.02199}, 2016.

\bibitem{long2015learning}
Mingsheng Long, Yue Cao, Jianmin Wang, and Michael~I Jordan.
\newblock Learning transferable features with deep adaptation networks.
\newblock {\em arXiv preprint arXiv:1502.02791}, 2015.

\bibitem{dabook09}
Joaquin Quionero-Candela, Masashi Sugiyama, Anton Schwaighofer, and Neil~D.
  Lawrence.
\newblock {\em Dataset Shift in Machine Learning}.
\newblock The MIT Press, 2009.

\bibitem{tobin2017domain}
Josh Tobin, Rachel Fong, Alex Ray, Jonas Schneider, Wojciech Zaremba, and
  Pieter Abbeel.
\newblock Domain randomization for transferring deep neural networks from
  simulation to the real world.
\newblock In {\em Intelligent Robots and Systems (IROS), 2017 IEEE/RSJ
  International Conference on}, pages 23--30. IEEE, 2017.

\bibitem{tzeng2015adapting}
Eric Tzeng, Coline Devin, Judy Hoffman, Chelsea Finn, Pieter Abbeel, Sergey
  Levine, Kate Saenko, and Trevor Darrell.
\newblock Adapting deep visuomotor representations with weak pairwise
  constraints.
\newblock {\em arXiv preprint arXiv:1511.07111}, 2015.

\bibitem{pmlr-v78-viereck17a}
Ulrich Viereck, Andreas Pas, Kate Saenko, and Robert Platt.
\newblock Learning a visuomotor controller for real world robotic grasping
  using simulated depth images.
\newblock In {\em Conference on Robot Learning (CoRL)}, 2017.

\end{thebibliography}

\end{document}